\pdfoutput=1
\documentclass[11pt,a4paper]{article}

\usepackage[T1]{fontenc}
\usepackage[utf8]{inputenc}
\usepackage[margin=1in]{geometry}
\usepackage{microtype}
\usepackage{booktabs}
\usepackage{array}
\usepackage{amsmath}
\usepackage{listings}
\usepackage{xcolor}
\usepackage[section]{placeins}
\usepackage{tikz}
\usetikzlibrary{arrows.meta,positioning,fit,backgrounds,calc}
\usepackage[breaklinks]{hyperref}
\usepackage{cleveref}

\definecolor{codebg}{gray}{0.96}
\lstdefinestyle{kah}{basicstyle=\ttfamily\small,backgroundcolor=\color{codebg},
  breaklines=true,columns=fullflexible,frame=none,xleftmargin=1em,
  showstringspaces=false}
\definecolor{kBlue}{HTML}{1F4E79}   \definecolor{kBlueL}{HTML}{DCE9F5}
\definecolor{kTeal}{HTML}{0F6F63}   \definecolor{kTealL}{HTML}{D8EEEA}
\definecolor{kAmber}{HTML}{A35B00}  \definecolor{kAmberL}{HTML}{FAE8CE}
\definecolor{kGrey}{HTML}{5A5A5A}   \definecolor{kGreyL}{HTML}{F0F0F0}

\hypersetup{colorlinks=true, linkcolor=kBlue, citecolor=kBlue,
            urlcolor=kBlue, filecolor=kBlue}

\tikzset{
  box/.style  = {draw, rounded corners=2pt, align=center, inner sep=5pt, font=\small},
  op/.style   = {box, draw=kGrey, fill=kGreyL},
  enf/.style  = {box, draw=kBlue, fill=kBlueL, thick},
  idn/.style  = {box, draw=kTeal, fill=kTealL},
  sink/.style = {box, draw=kAmber, fill=kAmberL, very thick},
  gap/.style  = {box, draw=kGrey, fill=white, dashed},
  lbl/.style  = {font=\scriptsize\itshape, align=center, text=kGrey},
  ar/.style   = {-{Stealth[length=5pt]}, thick, draw=kGrey},
  arA/.style  = {-{Stealth[length=5pt]}, thick, draw=kAmber},
  bnd/.style  = {draw, dashed, rounded corners=4pt, inner sep=10pt}
}

\newcommand{\prim}[1]{\textsc{#1}}

\title{\textbf{Five Primitives for Governing Autonomous AI Agents at Runtime}}

\author{%
  Jiten Oswal \qquad John Cadeddu\\[5pt]
  \normalsize Aurite AI\\[3pt]
  \normalsize \texttt{jiten@aurite.ai} \qquad \texttt{john@aurite.ai}\\[1pt]
  \normalsize \footnotesize ORCID (J.~Oswal):
    \href{https://orcid.org/0009-0009-8866-869X}{0009-0009-8866-869X}%
}
\date{August 25, 2026}

\begin{document}
\maketitle

\begin{abstract}
\noindent
Enterprise deployments of autonomous AI agents inherit a control model built for
human users and long-lived services, and the fit fails in three specific ways:
agent principals are \emph{ephemeral}, appearing and vanishing faster than
provisioning; their actions are \emph{selected by a model} rather than
programmed, so the set of things they may attempt is not known in advance; and
the population is \emph{discovered} rather than provisioned, because anyone who
can call an API can create one. We argue that governing such agents is a runtime
problem --- not a model-alignment problem and not a build-time problem --- and we
derive five primitives from the questions that must be answered before an action
takes effect and after it has: \prim{discovery}, \prim{identity},
\prim{governance}, \prim{attestation}, and \prim{supply chain}. For each we state
what fails if it is absent and why the others cannot structurally supply it. We
describe an implementation in which an agent's action is mediated against policy
before it takes effect, authorised against a per-tenant action vocabulary, and
recorded in a hash-linked signed ledger a third party can verify with the vendor
out of the loop. We report what the architecture costs: the enforcement point
sits on the request's critical path, identity requires a sidecar per workload,
and fail-closed mediation converts availability incidents into denial. We are
explicit about implementation status: four primitives are built and running in
private pilots, and the fifth is built as separate tooling and not yet
integrated into the request path. We keep it in the set
deliberately: a five-part decomposition that exactly matches what its authors
happened to build is not a taxonomy but a description of a codebase.
\end{abstract}

\section{Introduction}\label{sec:intro}

An enterprise deploying an autonomous agent is deploying a principal that acts
on its behalf, against its systems, using its credentials. Every mature
organisation already has machinery for that situation --- identity providers,
role-based access control, API gateways, audit logs --- built over three decades
for two kinds of principal: human users, who are onboarded, and services, which
are provisioned.

An agent is neither, and the mismatch is not a matter of degree.

\paragraph{Agents are ephemeral.} A service is provisioned once and runs for
months; a human is onboarded once and stays for years. An agent instance may
exist for the duration of a single task. Any control that depends on a
registration step performed by a person before the principal acts will either be
skipped or will become the bottleneck that causes the agent to be run outside
the control.

\paragraph{Agents choose their actions.} A service's call sites are enumerable
by reading its code. An agent's are not: the action is selected at runtime by a
model, from a tool set, with parameters the model composes. Orchestration
frameworks of the kind now in common use~\cite{auriteframework} make this the
default shape (an agent is handed a set of tools and decides among them), so the
assumption that the set of governable operations is known when the policy is
written does not survive. An agent will attempt things nobody anticipated. That
is what it is for.

\paragraph{Agents are discovered, not provisioned.} Anyone who can call an API
can create one. In practice the population inside an enterprise is not a list
somebody maintains; it is a fact to be measured. A control plane that governs
only the agents it was told about governs the compliant subset, which is not the
interesting one.

\paragraph{The claim.} These three properties make agent governance a
\emph{runtime} problem. It is not primarily a model-alignment problem: an agent
that has been perfectly aligned still needs its authority bounded, because
alignment is a property of the model and authority is a property of the
deployment. It is not primarily a build-time problem either: static analysis can
constrain what an agent's code may do, but not what a model chooses to do with
parameters it composes at the moment of the call.

This paper presents a decomposition of that runtime problem into five
primitives, derived in \cref{sec:derivation} from the questions that must be
answered before an action takes effect and after it has:

\begin{enumerate}\itemsep2pt
  \item \prim{Discovery}: \emph{does this agent exist, and do we know about it?}
  \item \prim{Identity}: \emph{what is it, provably?}
  \item \prim{Governance}: \emph{may it do this, decided before it does?}
  \item \prim{Attestation}: \emph{what happened, in a form someone else can check?}
  \item \prim{Supply chain}: \emph{what is it made of, and has that changed?}
\end{enumerate}

\paragraph{Contributions.}
\begin{enumerate}\itemsep2pt
  \item An argument that these five are \emph{irreducible} for this problem
        (\cref{sec:derivation}): for each, what fails in its absence and why no
        other primitive can supply it.
  \item Design consequences that are \emph{specific to agent principals}
        (\cref{sec:discovery}--\cref{sec:attestation}) and would not arise for
        human or service principals, including why a discovery surface for
        agents must aggregate rather than append, and why it must be
        structurally incapable of storing a value.
  \item A report on \textbf{governing the governor}
        (\cref{sec:selfgov}): the control plane's own agents run through the
        same identity, mediation, kill switch and attestation path, in a
        separate policy namespace enforced at the transport layer rather than by
        naming convention.
  \item An honest account of what the architecture costs
        (\cref{sec:cost}): critical-path latency, a sidecar per workload, and
        the conversion of availability incidents into denials.
  \item A status report (\cref{sec:status}) distinguishing what is built and
        running from what is built but not yet integrated, and an argument for
        why that primitive belongs in the set regardless.
\end{enumerate}

\Cref{app:trace} traces one action end to end through the four built
primitives, and names what the fifth would have added.

\paragraph{What this paper is not.} It is not an evaluation. We report
implementation status and design rationale, not measured outcomes across a
deployment population. \Cref{sec:status} says why we are not in a position to
offer the latter, and what would be needed.

\section{Why the existing controls do not fit}\label{sec:existing}

The existing machinery fails in specific ways, and each failure has a
workaround that does not survive contact with the properties in
\cref{sec:intro}.

\paragraph{Human IAM.} Directory-based identity assumes a lifecycle anchored in
a person: joiners, movers, leavers. Agents have no such lifecycle, and the
common workaround --- give the agent a service account, or worse, a human's
credentials --- destroys attribution exactly where it matters. An action taken
by an agent using a person's credential is indistinguishable, in every
downstream log, from an action taken by that person.

\paragraph{Service identity.} Workload identity systems fit far better, and we
build on one (\cref{sec:identity}). What they do not supply is
\emph{authorisation semantics for probabilistic callers}: they answer \emph{which workload is this}, which is necessary and not sufficient when the caller decides
its own actions.

\paragraph{API gateways and network policy.} Both operate on the wrong
granularity. A gateway can permit or deny a route; the governance question for
an agent is usually about the \emph{parameters}: may it email this recipient,
move this amount, read this record class. Those live in the payload. A
network policy that permits an agent to reach the payments API has said nothing
about which payments it may make.

\paragraph{Application-level guardrails.} Checks written inside the agent's own
process are the most common approach in practice and the weakest structurally:
the enforcement point shares a failure domain with the thing it constrains, and
it is modified by the same team, in the same deployment, under the same
pressure. It also cannot produce evidence a third party will accept, for the
same reason.

\paragraph{Audit logs.} Conventional logging produces a record written,
stored and displayed by the same party whose behaviour it evidences. That is
adequate for debugging and inadequate for the case that actually matters, which
is a dispute.

The pattern across all five is that each mechanism is \emph{correct for the
principal it was designed for}. This paper does not argue that those systems are
wrong. It argues that a layer above them has to answer a set of questions none of
them was built to answer.

\subsection{What actually runs}\label{sec:arch}

\begin{figure}[!htb]
\centering
\begin{tikzpicture}[node distance=6mm]
  \node[op] (agent) {Agent\\+ sidecar};
  \node[enf, right=14mm of agent] (srv) {Management server\\\scriptsize mediator · attestation · verifier};
  \node[idn, above=9mm of srv] (idp) {Identity plane};
  \node[sink, below=20mm of srv] (db) {Chain (DB) + signing keys};
  \node[op, below=9mm of agent] (con) {Operator console};

  \draw[ar] (agent) -- node[lbl, above, text width=17mm]{mTLS\\authorize} (srv);
  \draw[ar, draw=kTeal] (srv) -- (idp);
  \draw[ar, dashed, draw=kTeal] (agent) |- (idp);
  \draw[arA] (srv) -- (db);
  \draw[ar] (con) -| ([xshift=-11mm]srv.south);

  \node[lbl, right=10mm of db, text width=34mm, anchor=west, text=kAmber]
    (ev) {evidence package $\rightarrow$ independent verifier, vendor out of the loop};
  \draw[arA, dashed] (db) -- (ev);
\end{tikzpicture}
\caption{The deployed shape. Colour carries meaning: blue is the enforcement
point, teal the identity plane and the credential path, amber the attested chain
and the export a customer verifies without us. The dashed path from the agent to
the identity plane is credential acquisition.}
\label{fig:arch}
\end{figure}
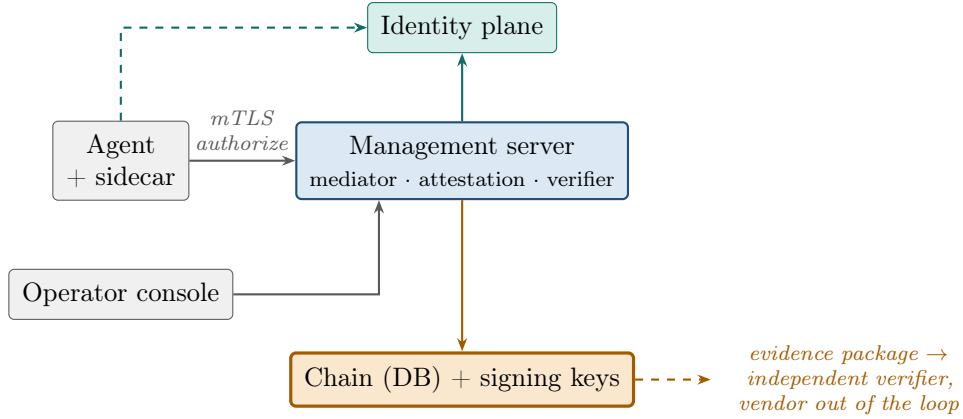

\Cref{fig:arch} shows the deployed shape, because the argument that follows is
easier to hold with the components named. An operator reaches a console through
an authenticating edge; the console is served by a gateway that fronts a
management API. The management server holds the mediator (policy evaluation and
the action catalog), the attestation factory and its background verifier, and
talks to an identity service that drives the workload-identity plane. State is a
relational database holding the append-only chain, and a key-management service
holding per-tenant signing keys that are never exported.

The wire contract for a governed agent is a single call ---
\texttt{POST /v1/actions/authorize} over mutual TLS, presenting a workload
credential --- and everything in \cref{sec:derivation} hangs off it.

\section{Deriving the five}\label{sec:derivation}

\begin{figure}[t]
\centering
\begin{tikzpicture}[node distance=7mm]
  \node[op] (d) {\prim{Discovery}\\\scriptsize does it exist?};
  \node[idn, right=9mm of d] (i) {\prim{Identity}\\\scriptsize what is it?};
  \node[enf, right=9mm of i] (g) {\prim{Governance}\\\scriptsize may it?};
  \node[sink, right=9mm of g] (a) {\prim{Attestation}\\\scriptsize what happened?};
  \draw[ar] (d)--(i); \draw[ar] (i)--(g); \draw[ar] (g)--(a);

  \node[gap, below=13mm of i, xshift=14mm, minimum width=62mm]
       (s) {\prim{Supply chain} --- \emph{what is it made of?}};
  \draw[ar, dashed] (s.north) -- ++(0,4mm) -| (i.south);
  \draw[ar, dashed] (s.north) -- ++(0,4mm) -| (g.south);

  \node[lbl, below=1.5mm of s] {a different timescale: provenance established
    before the request, invalidated between requests};
  \node[lbl, above=1mm of g] {request time};
\end{tikzpicture}
\caption{The four request-time questions, and the fifth that operates between
requests. Colours carry the same meaning as \cref{fig:arch}: teal is identity,
blue is the enforcement point, amber is the attested record. Supply chain is
drawn detached because it is not a step in the pipeline: it conditions whether
the answers the other primitives give are worth anything.}
\label{fig:five}
\end{figure}
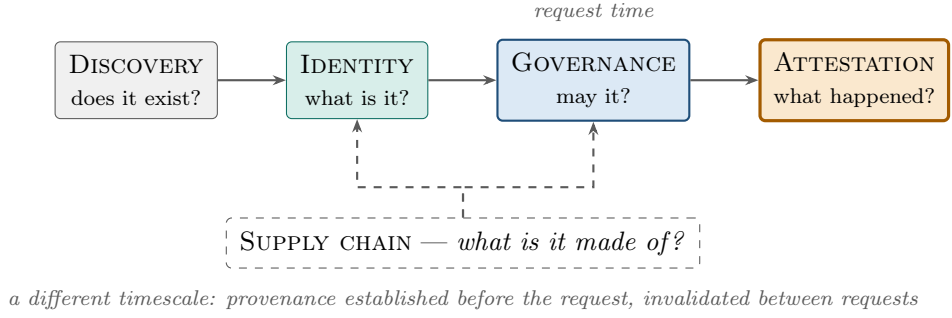

Consider a single agent action --- a payment, an email, a record write --- and
ask what must be true for an enterprise to permit it and, afterwards, to be able to
prove what occurred. Four questions arise in sequence, and a fifth cuts
across all of them on a different timescale (\cref{fig:five}).

\paragraph{1. Does this agent exist, and do we know about it?} If the population
is discovered rather than provisioned (\cref{sec:intro}), then every other
control applies only to the subset already known. \emph{No other primitive
supplies this}: identity can authenticate an agent that presents itself,
governance can decide about an agent that asks, attestation can record an agent
that acted, and all three are triggered by the agent's own participation. An agent that never
presents itself is invisible to all of them, and it is precisely the
ungoverned agent that will not present itself.

\paragraph{2. What is it, provably?} A policy decision needs a subject, and the
subject must be established by something stronger than an assertion in a header.
\emph{Discovery cannot supply this}: knowing a population exists is not the
same as authenticating a member of it at the moment it acts.

\paragraph{3. May it do this --- decided before it does?} An authorisation
decision must precede the effect, or it is not a control but a report.
\emph{Identity cannot supply this}: a proven identity with no policy attached
is a label. This is also where the agent-specific difficulty concentrates,
because the action set is not known when the policy is written
(\cref{sec:governance}).

\paragraph{4. What happened, in a form someone else can check?} Every answer
above is produced by the governance layer itself. Without an independently
verifiable record, an enterprise asking \emph{did your system actually stop that?}
receives an answer from the party being asked. \emph{Governance cannot supply
this}: a decision engine attesting to its own decisions is exactly the structure
whose trustworthiness is in question.

\paragraph{5. What is it made of, and has that changed?} The first four
primitives, working perfectly, will faithfully identify, authorise and record
the actions of an agent whose dependency was replaced yesterday with a
compromised version. This question does not sit in the request path: it is
answered before the request and can be invalidated between requests, which is
why it is drawn detached in \cref{fig:five}. \emph{None of the other four can
supply it}: they operate on the agent's \emph{behaviour}, and this is a question
about its \emph{composition}.

\subsection{Why not fewer, and why not more}

The set is minimal in the sense that removing any one leaves a failure the
remaining four cannot address, as argued above. It is not maximal: we make no
claim that no sixth primitive exists. What we claim is that these five are
each necessary, that they are separable --- each can be implemented, reasoned
about and failed independently --- and that the separation is load-bearing rather
than presentational, because the primitives have genuinely different trust
models, failure modes and timescales.

The most important consequence of the separation is the one in \cref{sec:cost}:
these primitives fail differently, and a design that merges them inherits the
worst failure mode of each.

\section{Discovery}\label{sec:discovery}

Discovery answers \emph{which agents exist and what are they attempting}. It has two
halves that behave very differently, and our experience is that the second is
both easier and more valuable than we expected.

\subsection{Discovering what agents attempt}

The first half is population discovery: enumerate the agents. The second is
\emph{capability} discovery: an agent attempts an action the platform has no
definition for, and that attempt is itself the signal.

This falls out of a governance layer that is deny-by-default. An action nobody
has defined cannot be permitted, but it can be \emph{recorded}, attributed,
and surfaced to an operator who then decides whether to define it. The refusal
and the discovery are the same event. An agent doing something unanticipated
stops being purely a failure and becomes the mechanism by which the policy
catches up with reality.

Three rules govern that recording surface. All three are consequences of the
principal being an agent; none would arise for a human user.

\paragraph{Aggregate, never append.} One row per (organisation, action type),
upserted, not one row per attempt. The reasoning is specific to the
principal: \emph{an agent meeting an action the platform cannot govern is
precisely the agent that retries hardest.} A per-attempt table would be
unbounded and driven by the very failure it records: a table an agent in a retry
loop can fill. A human user, refused, reads the error and stops. This is the
general shape of a class of agent-specific hazard, a failure signal that an agent
responds to by amplifying, which turns a diagnostic into a denial of service
against the diagnostic.

\paragraph{Names, never values.} An unregistered action is, by definition, one
whose fields nobody has classified. Nothing is known about which of its values
are safe to retain, so the correct answer is to retain none. Only parameter
\emph{names} are stored, and the table has no column a value could go in, so the
unsafe state is unrepresentable rather than merely avoided. This is
the disclosure discipline of the governance layer~\cite{c2minimization} extended
to the discovery surface, and it matters most here: discovery is exactly where
the platform is dealing with data it has no classification for.

\paragraph{Recording is observability, not governance.} The action is already
being refused by the time the recorder runs, so a failure in recording must not
change that outcome or surface to the caller. The worst case is that discovery
loses a data point, which is strictly better than a recording bug becoming an
outage on the authorisation path. A discovery mechanism must not be able to
deny, or it acquires the blast radius of the thing it observes.

\subsection{Discovering the agents themselves}

The other half --- enumerating agents by scanning code, deployment manifests and
network behaviour --- is the part we have implemented as a library and not yet
placed in the running system (\cref{sec:status}). We include the design
reasoning because the requirement is real regardless of our status.

The hard part is not finding candidates; it is \emph{reconciliation}. Scanning
produces claims about agents from several sources of differing reliability:
code that constructs an agent, a deployment that runs one, a credential that was
issued to one. These do not agree. An agent in code that is never deployed
is not a governance problem. A workload holding a credential that appears in no
code is a much larger one. The output that matters is therefore not a list but a
\emph{disagreement}: the set of things visible to one source and not another,
which is where ungoverned agents are found.

\section{Identity}\label{sec:identity}

Identity answers \emph{what is this, provably}. For an ephemeral principal the
proof cannot depend on anything a person did beforehand.

We use attestation-based workload identity~\cite{spiffe}: the platform observes
properties of the running workload it cannot choose for itself --- what it is,
where it runs, under which service account --- and issues a short-lived
credential on that basis. The agent presents it over mutual TLS. Nothing is
provisioned by hand, credentials expire on the order of minutes, and revocation
is a matter of not reissuing.

Three consequences are where the design differs from conventional service
identity.

\paragraph{The identity is presented at the transport layer, not asserted in the
payload.} The identity a policy evaluates is the one the connection proved, not
one the agent claimed. This sounds obvious and is easy to lose: an agent
framework that forwards a caller identity in a header, and a mediator that reads
it, together produce a system where the agent chooses its own principal.

\paragraph{One principal needs several vocabularies.} A single agent identity is
expressed differently in storage, in the audit envelope, and as a policy entity,
because each layer has different constraints. What makes this tractable is a
canonical mapping between them, so one identity can be traced across all three
during an incident. What makes it necessary is that collapsing them couples the
policy schema to the database schema and the audit format to both, and the
audit format is the one that must not change, because entries are permanent.

\paragraph{Some principals are better left underivable from storage.}
Platform-internal agents are per-tenant instances of a small number of classes.
Storing a row per (tenant, class) buys nothing: the identifier is derivable from
the pair, and the operational state that actually varies lives elsewhere. The
general rule we would apply again: store identity only where identity carries
information; where it is a deterministic function of things already
stored, derive it.

\section{Governance}\label{sec:governance}

Governance answers \emph{may it}, before the action takes effect. The mediator sits
on the request path: the agent asks, a policy engine decides, and the agent
proceeds or does not.

Two things about this are different for agents.

\subsection{The action vocabulary cannot be fixed in advance}

If the set of governable action types is a fixed enumeration shipped by the
vendor, then an agent that refunds an order or provisions a tenant is not
partially governed: it is \emph{ungovernable}, and the customer waits for a
release. Since an agent's actions are model-selected from a tool set the
customer defines, the vendor cannot enumerate them even in principle.

So the action vocabulary is per-tenant and customer-authored, and the discovery
surface of \cref{sec:discovery} is what tells an operator which entries are
missing. The loop is: attempt $\rightarrow$ refused $\rightarrow$ visible
$\rightarrow$ defined $\rightarrow$ governable.

The cost of extensibility is that old audit becomes harder to interpret: once
a definition can change, a record naming an action does not by itself say what
that action meant at the time. The answer is to make definitions immutably
versioned and to have each attested record point at the version that governed
it~\cite{c2minimization}.

\subsection{Deny-by-default protects authorisation and not disclosure}

Our mediator refuses any action no policy permits, which is a complete
protection for the authorisation question and \emph{no protection at all} for a
second question hiding inside the same configuration: how much of the action's
parameters cross the boundary in order for the policy to be evaluated.

That is a disclosure decision, it is governed by the same tables, and nothing
about deny-by-default constrains it: reclassifying a field to make it visible
to policy denies nothing and therefore signals nothing. We treat this at length
in a companion paper~\cite{c2minimization}; it is named here because the
mistake is invited by the architecture: two decisions, one control surface, and
only one of them has a safety net.

\subsection{Stopping an agent that is already running}

A control that can only refuse new requests is insufficient for a principal that
acts in a loop. The system carries a kill switch whose state is consulted on
every authorisation, so revocation takes effect on the next action rather than
at the next credential expiry. The design constraint worth recording is that
this check is on the hot path of every request, which puts a hard ceiling on how
expensive it may be, and that a kill switch which is slow is one operators
hesitate to use.

\section{Attestation}\label{sec:attestation}

Attestation answers \emph{what happened, in a form someone else can check}. It is the
primitive most often skipped, because from inside a system it is indistinguishable
from logging, and it is the one that determines whether any of the others can be
relied upon in a dispute.

Every authorisation decision is written to an append-only ledger as a row bound
by hash to its predecessors, canonicalised, and signed by a per-tenant key held
in a key-management service and never exported. The properties claimed are
position binding, authenticity, completeness and forgery resistance, argued from
stated cryptographic assumptions.

The part that matters for this paper is not the chain construction, which is
conventional~\cite{crosby2009,rfc6962}, but where the verification happens.

\paragraph{The verifier must run somewhere we do not.} A customer can export an
evidence package and verify it with an independent tool, with the vendor's
systems entirely out of the loop. This is the only configuration in which the
answer to \emph{did your system really do that} is worth anything, because every other
configuration routes the question through the party being asked.

\paragraph{Verification is a claim about authenticity \emph{and} completeness,
and the second is easy to get wrong.} A window of records that all verify
individually is not thereby the whole record. An export path must state what
window it covers and whether that window is complete, and a completeness field
that cannot determine the answer must resolve to \emph{false} rather than to a
reassuring default. That is a failure we made and fixed.

\paragraph{An honest residual.} The record is signed by infrastructure an
operator controls. An operator with both database and signing access could, in
principle, rewrite history and re-sign it; what the chain provides against that
is evidence of the rewrite to anyone holding a prior export, not prevention. We
state this because a trust model that names its residuals is usable and one that
does not is decoration.

\section{Supply chain}\label{sec:supplychain}

Supply chain answers \emph{what is this agent made of, and has that changed}. It is
the primitive in this set that does not yet run inside the control plane
(\cref{sec:status}), and this section argues why it stays in the set.

Unlike the other four, this primitive has begun to attract dedicated work:
bills of materials scoped to agentic systems, for
instance~\cite{agentriskbom}. What follows argues why the question belongs
\emph{in this set}, alongside the runtime primitives, rather than in a separate
discipline.

The first four primitives share an assumption so basic it is easy to miss: that
the thing being identified, authorised and recorded is the thing the operator
believes it to be. An agent is a composition --- a model, a framework, a tool
set, a dependency tree, a container image --- and every element is a supply-chain
surface. An agent whose dependency was replaced yesterday is, from the
perspective of the other four primitives, in perfect order: it authenticates
correctly, it requests only permitted actions, and every decision is faithfully
attested. The record will be an accurate account of a compromised thing
behaving legitimately.

Three properties make this worse for agents than for conventional software.

\paragraph{The trust decision is made once and consumed continuously.} A
provenance check at deploy time conditions every subsequent request. Between the
check and the request, the composition can change: a pulled tag, a
transitively updated dependency, a model endpoint that now routes elsewhere.

\paragraph{The composition includes things not conventionally in scope.} The
model is a dependency. So is the prompt template, and so is the tool
description the model reads to decide what to call. Modifying a tool description
changes agent behaviour without touching a line of code or a package version,
and no conventional supply-chain tool looks there.

\paragraph{Attribution across the boundary is unsolved.} When an agent takes a
harmful action, distinguishing \emph{the model chose badly} from \emph{the tool description was manipulated} from \emph{a dependency was compromised} requires provenance the
other four primitives do not collect.

\paragraph{Where the capability does exist.} The question is not one we have
ignored, and where it sits is worth stating precisely. Provenance for agent
composition is addressed today in separate developer tooling built by the same
organisation: static validation of agent code and its dependencies before it
ships~\cite{agentverifier}, and a dedicated composition-analysis product serving
several use cases, itself in private pilot. Neither runs inside the control
plane described here, and neither conditions an authorization decision.

The integration path is specific rather than aspirational. The build-time
analysis is not confined to build time by anything essential: what it computes
about an agent's composition can be evaluated against a running workload, and
the work is to expose a provenance verdict the mediator can consult on the
request path. That is what would close the gap \cref{sec:status} records, and it
is where the effort goes next. The design of that integration, and what it
implies for the request path's latency budget (\cref{sec:cost}), needs more room
than a subsection here; we defer it to a forthcoming companion paper on
provenance for agent composition.

We state this as a fact about our roadmap and not as an argument, and the
distinction matters: the primitive belongs in the set whether or not we ever
complete that integration, because the question it answers is one any
governance layer for agents has to answer. A reader should not read this
paragraph as a claim that the control plane described in \cref{sec:status}
enforces supply-chain provenance on the request path. It does not.

\paragraph{Why it stays in the set.} We could have proposed four primitives and
matched our implementation exactly. We think that would have been a worse
paper and a worse taxonomy. The set is derived in \cref{sec:derivation} from the
questions that must be answered, not from the modules we happen to have
shipped, and this is the question the control plane cannot yet answer for
itself. Removing it would make the decomposition tidier and would quietly
redefine the problem as the part we had wired up.

\section{The control plane governs its own agents}\label{sec:selfgov}

A governance layer is itself automation. It runs agents of its own --- one that
consumes attestations, one that delivers evidence to a customer's SIEM, one that
reconciles identity state --- and each performs privileged actions against
customer data. A reader entitled to be skeptical will ask whether the vendor's
own automation is subject to the controls the vendor sells, and for most
products the answer is no: the platform's internal jobs run with ambient
privilege, outside the mechanism.

Ours run through the same path. Kahuna's internal agents obtain workload
credentials from the same identity plane, call the same authorization endpoint,
are refused by the same deny-by-default mediator, are stopped by the same kill
switch, and emit into the same attestation chain. There is no privileged
bypass, and the framework they are built on exposes exactly the five hooks the
primitives imply: an identity provider, an action mediator, an attestation sink,
a kill-switch source, and a governance context that binds them.

Self-governance is not free, and two design consequences follow.

\paragraph{Internal actions need their own namespace, and naming is not
enough.} Internal action types live in a separate policy namespace from customer
action types and must never appear on customer-facing API surfaces. A separate
namespace prevents a customer policy from accidentally authorising an internal
action, and --- more importantly --- prevents an internal action from being
evaluated against a customer's policy set, which would make the platform's own
behaviour contingent on customer configuration.

But a namespace is a naming convention, and a naming convention is not an
enforcement boundary. The authorization route therefore checks, at the
transport layer, that the calling workload credential belongs to an
internal-agent identity before it will accept an internal action type at all.
The namespace expresses the separation; the identity check enforces it. This is
the same principle as \cref{sec:identity}'s transport-layer binding, applied to
the platform's own callers, and it is what stops the separation from being
decorative.

\paragraph{The audit envelope stays closed, and the distinction moves
elsewhere.} The attested record's principal-type vocabulary is a closed set, and
adding a value to it would be a schema change to an append-only ledger, a cost
out of all proportion to the benefit. So internal agents are recorded under an
existing principal type, with the additional forensic distinction carried in a
separate field. \textbf{A permanent record's schema should be the last thing a new feature is
allowed to change}, and the discipline that follows is to ask what can be added
beside the envelope before asking what can be added to it.

\paragraph{What self-governance is and is not evidence of.} It demonstrates
that the primitives are general enough to govern the governor, which is a real
architectural claim and the one we would make. It is not evidence that the
controls are correct: a flawed mechanism applied to oneself is still flawed.
Its value is that it removes an entire class of question about privileged
exceptions, and that we discover the ergonomic problems of our own integration
path before a customer does.

\section{What the architecture costs}\label{sec:cost}

A control plane on the request path is not free, and proposals of this kind are
usually presented without a price. Ours has four, and we would rather state them
than have a reader discover them during an evaluation.

\paragraph{Latency on the critical path.} An action mediated before it takes
effect means a network round trip, a policy evaluation and a durable write
between the agent deciding and the agent acting. There is a structural tension
here: extending a \emph{signed} chain synchronously with the decision serialises
requests behind the chain head, which inflates tail latency
while leaving the median merely poor. The remedy --- returning the decision
after policy evaluation and a durable enqueue, with signing completing under a
bounded lag --- is a change to \emph{when} the record becomes tamper-evident
relative to the action, and is therefore a security decision rather than a
performance one. It should be taken as such.

\paragraph{A sidecar per workload.} Attestation-based identity requires
something alongside the agent to obtain and rotate credentials. That is an extra
container, a shared volume, a registration entry, and a deployment change for
every governed agent. It is the single largest adoption obstacle we have
encountered, and it is intrinsic rather than incidental: identity that a
workload can assert for itself is not attested identity.

\paragraph{Fail-closed converts availability incidents into denials.} If the
mediator is unreachable, the correct behaviour for a security control is to
refuse. This means an outage in the governance layer becomes an outage in every
governed agent. We think this trade is right --- a governance layer that fails
open is a governance layer whose guarantees hold only when nothing is wrong,
which is the wrong time --- but it is a real operational cost and it moves the
availability requirement of the control plane up to match the most critical
thing it governs.

\paragraph{Integration effort is front-loaded.} Comparing two functionally
equivalent agents, one ungoverned and one mediated, the governed version
requires several times the code and additional deployment resources. The large
majority of that is reusable infrastructure, so the marginal cost of the tenth
agent is a small fraction of the first. Both numbers matter and they answer
different questions: the first determines whether anyone integrates at all, and
the second determines whether the programme scales.

\paragraph{When this is the wrong architecture.} If an agent's action set is
small, fixed, and enumerable in advance; if it runs in a single trust domain
with no external counterparties; and if no third party will ever need to be
convinced of what it did, then in-process guardrails are simpler, faster and
sufficient. The architecture here earns its cost when the action set is open,
the population is not enumerable, or the evidence has to satisfy someone who
does not trust the operator.

\section{Implementation status}\label{sec:status}

\begin{table}[h]
\centering\small
\begin{tabular}{@{}llp{74mm}@{}}
\toprule
\textbf{Primitive} & \textbf{Status} & \textbf{What that means precisely} \\
\midrule
\prim{Discovery}    & Partial & Capability discovery (\S\ref{sec:discovery}) is built, wired and running: unregistered attempts are recorded, attributed and surfaced. Agent-population discovery by code scanning exists as a tested library with \emph{no execution path}: nothing invokes it. \\
\prim{Identity}     & Built   & Attestation-based workload identity, mutual TLS, transport-layer binding, health-gated readiness. The supported integration is the sidecar; SDK-native credential acquisition is not implemented. \\
\prim{Governance}   & Built   & Policy evaluation on the request path, per-tenant action catalog, deny-by-default, kill switch consulted per request. \\
\prim{Attestation}  & Built   & Hash-linked signed chain, per-tenant keys, background verification, evidence export, independent offline verifier published. \\
\prim{Supply chain} & \textbf{Partial} & Composition analysis is built and in private pilot as separate developer tooling (\cref{sec:supplychain}); it is \emph{not} integrated into the control plane. In-repo sensors are scaffolds and nothing collects provenance on the request path today, pending the latency question of \cref{sec:cost}. \\
\bottomrule
\end{tabular}
\end{table}

\paragraph{Deployment scope.} The system runs in a small number of private
pilots rather than as a general release. This is deliberate: what a governance
layer must express is determined by what organisations actually need governed,
and that is learned from a few deployments studied closely rather than many
observed shallowly.

\paragraph{What we therefore claim, and do not.} We claim that the
decomposition in \cref{sec:derivation} is sound, that four of the five are
implemented and exercised, and that the design consequences reported in
\cref{sec:discovery}--\cref{sec:attestation} were arrived at by building the
thing and being corrected by it. We do \emph{not} claim measured outcomes across
a deployment population, evidence that the five primitives are sufficient for
organisations unlike our pilots, or any operational result about the primitive we
have not integrated. Those require a larger and more varied deployment base than we
have.

\paragraph{One methodological note, because it shaped several designs.}
Building this system produced a recurring failure in which a component reported
healthy while being structurally incapable of doing its job: a verifier that
verified nothing for fifty days while reporting ready, an alert catalogue with
nothing evaluating it, a readiness gate that by construction could never become
false. In each case the signal was derived from the \emph{absence of a negative}
rather than the presence of a positive, so a disconnected component and a
working one were indistinguishable at the reporting surface. That experience is
why \cref{sec:status} distinguishes \emph{built} from \emph{wired}: in this class
of system those are different states, and a status table that conflates them is
the first thing a reader should distrust.

\section{Related work}\label{sec:related}

\paragraph{Workload identity and zero trust.} Attestation-based workload
identity~\cite{spiffe} and the zero-trust architecture it
serves~\cite{nist800207} supply the identity primitive and the assumption that
network position confers no authority. Neither addresses authorisation semantics
for a caller that selects its own actions.

\paragraph{Policy engines.} Expressive, analysable authorisation
languages~\cite{cedar,opa} give the governance primitive its decision procedure.
What they assume, reasonably, is that the vocabulary of actions and resources is
known to whoever writes the policy, the assumption \cref{sec:governance}
argues does not hold for agents.

\paragraph{Tamper-evident logging and transparency.} Hash-linked logs and
transparency architectures~\cite{crosby2009,rfc6962} supply the attestation
primitive's construction. Our contribution is not the construction but the
insistence that verification run outside the vendor's systems.

\paragraph{Software supply-chain integrity.} Provenance frameworks and artifact
attestation~\cite{slsa,intoto} address composition for conventional software.
\Cref{sec:supplychain} argues that agents extend the relevant surface to models,
prompt templates and tool descriptions, which those frameworks do not currently
cover.

\paragraph{Agent safety and alignment.} A large literature addresses making
models behave well. This paper is orthogonal to it: we assume a model that may
be well-aligned and still deployed with more authority than intended, and we
constrain the deployment rather than the model. Build-time validation of agent
code~\cite{agentverifier} is the closest complement: it constrains what an
agent's code may do, where this constrains what a running agent may cause.

\paragraph{Agent identity.} Interest in identity for non-human and agentic
principals is active, and a recent survey maps the standards landscape and its
gaps~\cite{otsuka2026}; our diagnosis about ephemerality and scale is shared.
Our addition is that identity alone is insufficient for a principal whose
actions are not enumerable in advance, which is why identity is one primitive
here and not the whole architecture. Work is appearing on several of these
primitives individually; what we have not found is a treatment that argues the
\emph{set}: which questions must be answered, in what order, and why answering
four of them well still leaves a gap.

\section{Conclusion}

Governing autonomous agents is a runtime problem because the three properties
that make agents useful --- they are ephemeral, they choose their own actions,
and they proliferate without provisioning --- are precisely the properties that
defeat controls designed for principals that are onboarded or provisioned.

The decomposition we propose follows from asking what must be true before an
action takes effect and after it has: the agent must be \emph{known}, it must be
\emph{provably identified}, the action must be \emph{decided upon in advance},
the outcome must be \emph{independently checkable}, and the thing being governed
must be \emph{what it is believed to be}. Each is necessary, none supplies
another, and they fail differently enough that merging them inherits the worst
failure of each.

Four of the five run in the control plane. We have kept the fifth in the set
because the question it answers does not go away while its integration is
pending, and because a taxonomy trimmed to its authors' implementation is not a
taxonomy. The most
useful thing we can offer another team is not our architecture but the ordering:
answer \emph{does it exist}, \emph{what is it}, \emph{may it}, \emph{what happened} and \emph{what is it made of}, then notice which of those your current controls
cannot answer at all.

\appendix
\section{One action, through four primitives}\label{app:trace}

The paper argues structurally throughout. This traces a single governed action
so the sequence is concrete. The agent is a customer's support agent; the action
is emailing a summary to a counterparty.

\begin{enumerate}\itemsep3pt

\item[\textbf{0.}] \textbf{Before any request: identity.} The agent's pod
starts. A sidecar obtains a short-lived workload credential from the identity
plane, issued on attested platform properties the workload cannot choose for
itself. Nothing was provisioned by a person, and the credential expires in
minutes.

\item[\textbf{1.}] \textbf{The agent decides.} A model selects
\texttt{send\_email} and composes its parameters: recipient, subject, body.
Note what has just happened: \emph{this} action with \emph{these} parameters did
not exist a moment ago, and no policy author could have enumerated it.

\item[\textbf{2.}] \textbf{Authorization: governance.} The agent calls
\texttt{POST /v1/actions/authorize} over mutual TLS. The identity the mediator
evaluates is the one the connection proved, not one the agent asserted
(\cref{sec:identity}). Cedar evaluates the tenant's policy against the action
type and its attributes; the kill-switch state is consulted on this same
request, so a revoked agent is stopped here rather than at credential expiry.

\item[\textbf{3a.}] \textbf{If the action type is unknown: discovery.} The
action is refused, because deny-by-default. It is also \emph{recorded}: one
upserted row per (tenant, action type), parameter names only, no column in which
a value could be stored (\cref{sec:discovery}). An operator sees that their
agents are attempting something the catalog does not define, and can define it.
The refusal and the discovery are the same event.

\item[\textbf{3b.}] \textbf{If the action type is known.} The decision ---
permit or deny --- returns to the agent, which proceeds or does not. The
decision preceded the effect, which is what makes this a control rather than a
report.

\item[\textbf{4.}] \textbf{Attestation.} The decision is written to the
append-only chain: hash-linked to its predecessor, canonicalised, and signed by
the tenant's key inside the key-management service. A background verifier
re-checks signatures and linkage on a cadence.

\item[\textbf{5.}] \textbf{Later: external verification.} The customer
exports an evidence package and runs an independent verifier. The vendor's
systems are not involved, which is the only configuration in which the answer to
\emph{did their system really do that} means anything (\cref{sec:attestation}).

\item[\textbf{--}] \textbf{What is missing.} Nothing in this sequence
establishes what the agent was \emph{made of} --- which model, which tool
descriptions, which dependency versions were in force at step~1. Every step
above would run identically for an agent whose dependency was replaced
yesterday. That is the gap \cref{sec:supplychain} names and that we have not
closed.

\end{enumerate}


\end{document}